# From Pixels to Damage Severity: Estimating Earthquake Impacts Using Semantic Segmentation of Social Media Images

Danrong Zhang[1], Huili Huang[2], N. Simrill Smith[3], Nimisha Roy[4], J. David Frost[5]

[1]Postdoctral Researcher, School of Computational Science and Engineering, Georgia Institute of Technology, Atlanta, GA, USA, 30332; email: dzhang373@gatech.edu;
[2]Graduate Research Assistant, School of Computational Science and Engineering, Georgia Institute of Technology, Atlanta, GA, USA, 30332; email: hhuang413@gatech.edu
[3]Water Engineer, Arup. Former Graduate Research Assistant, School of Civil and Environmental Engineering, Georgia Institute of Technology, Atlanta, GA, USA, 30332; email: simrill.smith@arup.com
[4]Lecturer, School of Computing Instruction, Georgia Institute of Technology, Atlanta, GA, USA, 30332; email: nroy9@gatech.edu
[5]Professor, School of Civil and Environmental Engineering, Georgia Institute of Technology, Atlanta, GA, USA, 30332; email: david.frost@ce.gatech.edu;

## Abstract

In the aftermath of earthquakes, social media images have become a crucial resource for disaster reconnaissance, providing immediate insights into the extent of damage. Traditional approaches to damage severity assessment in post-earthquake social media images often rely on classification methods, which are inherently subjective and incapable of accounting for the varying extents of damage within an image. Addressing these limitations, this study proposes a novel approach by framing damage severity assessment as a semantic segmentation problem, aiming for a more objective analysis of damage in earthquake-affected areas. The methodology involves the construction of a segmented damage severity dataset, categorizing damage into three degrees: undamaged structures, damaged structures, and debris. Utilizing this dataset, the study fine-tunes a SegFormer model to generate damage severity segmentations for post-earthquake social media images. Furthermore, a new damage severity scoring system is introduced, quantifying damage by considering the varying degrees of damage across different areas within images, adjusted for depth estimation. The application of this approach allows for the quantification of damage severity in social media images in a more objective and comprehensive manner. By providing a nuanced understanding of damage, this study enhances the ability to offer precise guidance to disaster reconnaissance teams, facilitating more effective and targeted response efforts in the aftermath of earthquakes.

# 1 Introduction

Earthquakes inflict significant losses on human society every year, making rapid response efforts crucial. In recent times, social media has emerged as a pivotal tool in disaster response due to its real-time capabilities and ability to provide detailed on-site information from eyewitnesses [1, 2]. In contrast with satellite or drone based imagery, social media typically bypasses the need for costly equipment, intricate data processing, and favorable weather [3]. When analyzed through advanced deep learning techniques, the imagery shared on social media platforms provides valuable insights into the extent of the damage [4]. Such information is invaluable, not only in assessing the severity of the situation but also in guiding the strategic deployment of disaster rescue, recovery and reconnaissance teams.

A number of established earthquake damage classification systems exist, such as the European Macroseismic Scale 1998 (EMS-98) [5], the Red Cross's damage assessment framework [6] and HAZUS [7], however, their applicability to social media imagery is limited. Table 1 provides a comparative analysis of various damage classification standards, aligning them on a scale from 0 to 10 to represent damage severity. In this scale, 0 indicates no damage, and 10 indicates complete destruction. It is evident that standards like EMS-98, Red Cross, and HAZUS are designed for on-site, expert-led assessments focusing on individual structures, necessitating detailed observations that trained professionals can provide. In contrast, social media images typically present scenes rather than specific structures. These scenes may encompass multiple buildings, or only portions of a building, and the observable damage can often be challenging to discern, as the assessment is confined to what is visible within the image. Consequently, the complexity and limited perspective inherent in social media imagery render these established engineering and disaster reconnaissance standards less effective for this context. Thus, while these standards are invaluable in traditional assessment scenarios, their direct application to the unique challenges of social media-based assessments is less feasible.

Numerous studies have approached the assessment of damage severity in social media images as a classification task, focusing on dataset development. The Damage Severity Assessment Dataset (DAD) categorizes damage into three levels: little-to-no, mild and severe damage, as shown in Table 1 [4]. Building on this, CrisisMMD extends the DAD classification system into categories of little-to-no, mild, severe, and indeterminate cases, labeled as "don't know or can't judge" [8]. Similarly, Alam et al. (2023) introduced MEDIC dataset, which adheres to the DAD standards for classifying damage severity in its tasks [9]. As Nguyen et al. (2017) highlight, terms like "severe", "mild", and "little" are inherently subjective, leading to variability in interpretation among different annotators, even with detailed descriptions provided [4]. Moreover, there is inconsistency in how annotators balance between the scope of damage and the degree of damage. For example, in instances where an image depicts a large area of mild damage versus a smaller area of severe damage, which scenario should be deemed more critically damaged?

To address the subjectivity and scope challenges in assessing damage severity from social media images, redefining the problem as semantic segmentation rather than a classification task could be the solution. Semantic segmentation offers a more granular analysis of images compared to classification, enabling the extraction of more accurate and detailed information regarding the extent of damage [10].

Significant research has been conducted on semantic segmentation for damage severity assessment from satellite and aerial imagery [11-17]. In contrast, applying semantic segmentation to social media images for damage detection remains relatively unexplored. Li et al. (2018) utilized a class activation map to identify damage-contributing areas in images [3]. Two annotators were asked to label the damage area of 10 images to create the ground truth and an Intersection-Over-Union (IoU) of 0.517 was reached. Zhang et al. (2020) defined visual attention as the area of an image that the artificial intelligence (AI) model will focus on to identify the damage severity [18]. They created a human-AI framework to have individuals annotating the visual attention area and an IoU of 0.598 was reached. Shekarizadeh et al. (2022) developed the deep-disaster guided back-propagation method for localizing damage in images from the DAD dataset, using visual analysis for evaluation due to the absence of ground truth in the dataset [2]. These studies underscore the need for a comprehensively annotated semantic segmentation dataset to provide ground truth and establish a baseline for model performance in damage severity assessment of social media images. The challenge also lies in the subjectivity in dataset construction, as the interpretation of concepts like 'damaged area' or 'visual attention' can vary among annotators.

This study addresses the challenge of assessing damage severity of social media images through a semantic segmentation approach. This method offers a more detailed, objective analysis and considers the extent of the damage in different areas of an image. Due to limited resources for labeling, the focus is narrowed to images of earthquake damage sourced from social media, rather than encompassing various disaster types. A specialized dataset of 547 images has been compiled by two annotators. This dataset includes three damage degree segmentations in each image: undamaged structures, damaged structures, and debris. A SegFormer model fine-tuned specifically for this social media derived earthquake dataset, demonstrates comparable, if not superior, accuracy to human annotators [21]. Furthermore, this research introduces a novel damage severity equation, which incorporates both segmentation and depth estimation within an image. The equation is designed to quantify the severity of damage in post-earthquake social media images more accurately and objectively.

Table 1. Comparative Analysis of Damage Classification Standards on a 0-10 Severity Scale

| Damage classification standard | 0 | 1 | 2 | 3 | 4 | 5 | 6 | 7 | 8 | 9 | 10 |
|---|---|---|---|---|---|---|---|---|---|---|---|
| EMS-98: Classification of damage to masonry buildings [5] | Negligible to slight damage | | Moderate damage | | Substantial to heavy damage | | | Very heavy damage | | Destruction | |
| | "(No structural damage, slight non-structural damage) Hair-line cracks in very few walls. Fall of small pieces of plaster only." | | "(Slight structural damage, moderate non-structural damage) Cracks in many walls. Fall of fairly large pieces of plaster. Partial collapse of chimneys." | | "(Moderate structural damage, heavy non-structural damage) Large and extensive cracks in most walls. Roof tiles detach. Chimneys fracture at the roof line; failure of individual non-structural elements (partitions, gable walls)." | | | "(Heavy structural damage, very heavy non-structural damage) Serious failure of walls; partial structural failure of roofs and floors." | | "(very heavy structural damage) Total or near-total collapse." | |
| HAZUS: Reinforced Masonry Bearing Walls with Wood or Metal Deck Diaphragms [7] | Not classified | Slight Structural Damage | | Moderate Structural Damage | | Extensive Structural Damage | | | | Complete Structural Damage | |
| | N/A | "Diagonal hairline cracks on masonry wall surfaces; larger cracks around door and window openings in walls with large proportion of openings; minor separation of walls from the floor and roof diaphragms" | | "Most wall surfaces exhibit diagonal cracks; some of the shear walls have exceeded their yield capacities indicated by larger diagonal cracks. Some walls may have visibly pulled away from the roof." | | "In buildings with relatively large area of wall openings most shear walls have exceeded their yield capacities and some of the walls have exceeded their ultimate capacities indicated by large, through-the-wall diagonal cracks and visibly buckled wall reinforcement. The plywood diaphragms may exhibit cracking and separation along plywood joints. Partial collapse of the roof may result from failure of the wall-to-diaphragm anchorages or the connections of beams to walls." | | | | "Structure has collapsed or is in imminent danger of collapse due to failure of the wall anchorages or due to failure of the wall panels." | |

| Red Cross[6] | | Af-fected | Minor | Major | De-stroyed |
|---|---|---|---|---|---|
| | N/A | "Mini-mal dam-age to the exterior and/or non-es-sential base-ments" | "Damage that does not affect structural integrity of the resi-dence." | "Residence sustained significant structural damages, re-quires extensive repairs." | "The res-idence is a total loss, or damaged to such an ex-tent that repair is not feasi-ble." |
| Damage Assessment Dataset (DAD) [4] | Little or no damage | | Mild damage | Severe damage | |
| | "Images that show damage-free infra-structure (except for wear and tear due to age or disrepair) be-long to the no-damage category." | | "Damage generally exceeding minor damage with up to 50% of a building, for example, in the focus of the image sustaining partial loss of amenity/roof. Maybe only part of the building has to be closed down, but other parts can still be used. In case of a bridge, if the bridge can still be used, but part of it is unusable" and/or needs some amount of repairs." | "Images that show substantial destruction of an infrastructure belong to the severe damage category. A non-livable or non-usable building, a non-crossable bridge, or a non-drivable road are all examples of severely damaged infrastructures." | |

## 2 Methodology

In this study, the SegFormer algorithm is fine-tuned for the segmentation of undamaged structures, damaged structures, debris and background in the images [21]. The IoU metric is employed to assess the fine-tuned SegFormer model's performance. This metric also serves to comparatively evaluate the annotations produced by two expert annotators. Additionally, the Dense Prediction Transformer (DPT) plays a crucial role in generating depth estimation maps for each image [19]. Integrating the segmentation results with the depth maps, a novel equation is proposed to calculate a damage severity score. This innovative approach allows for a nuanced consideration of varying degrees of damage and the area of the affected part in the images. Figure 1 shows a general workflow of the steps taken in this study. The subsequent sections will summarize the specifics of IoU, SegFormer, DPT, and the damage score equation.

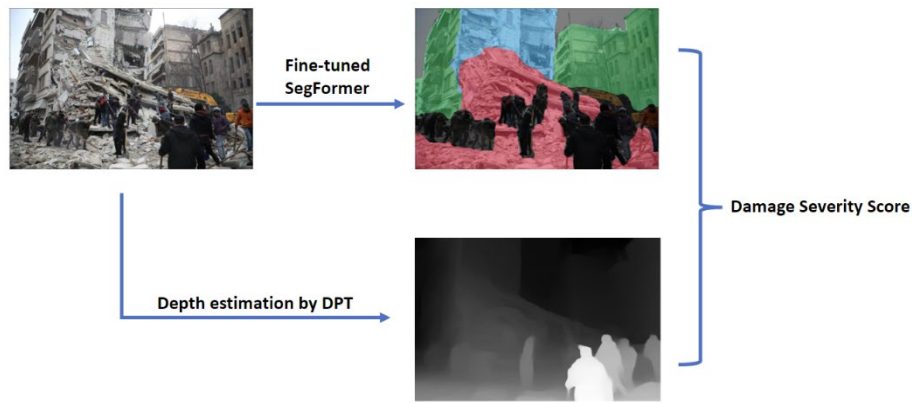

Fig. 1. Damage severity assessment workflow

### 2.1 IoU

IoU is a widely recognized metric for gauging the similarity between two annotations. It quantifies the extent of overlap between them. The method for calculating IoU is illustrated in Figure 2, where the division of areas effectively represents the degree of overlap between the two annotations. This approach allows for a clear and measurable assessment of how closely the annotations correspond to each other. In a multiclass segmentation scenario, the IoU of each class will be calculated to yield the average value as the final IoU.

The range of IoU is between 0 to 1. Lee and Chen (2021) point out that in general, an IoU above 0.5 can be considered as good prediction [20]. Li et al. (2018) offers a nuanced perspective in the context of damage localization. They argue that in the damage localization problem, where the subject of interest is not a discrete object, a slightly lower IoU threshold of 0.4 may be more appropriate than the standard 0.5 [3].

Considering that no published semantic segmentation dataset focused explicitly on damage assessment exists, there is no commonly accepted threshold for IoU in this domain.

In this study, the final IoU for each image is determined by averaging the IoU scores across four categories, which are the three damage degrees, namely undamaged structure, damaged structure, and debris, plus the background. Figure 3 presents the comparison between the annotation mask and the segmentation result obtained from the finely-tuned SegFormer, illustrating undamaged structures in green, damaged structures in blue, debris in red, and the remaining areas as background. The IoU between those two images is 0.69.

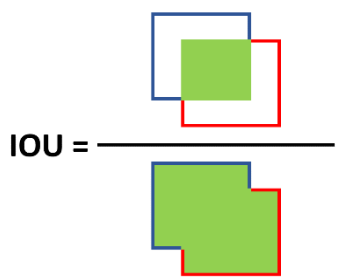

**IOU =**

Fig. 2. IoU metric

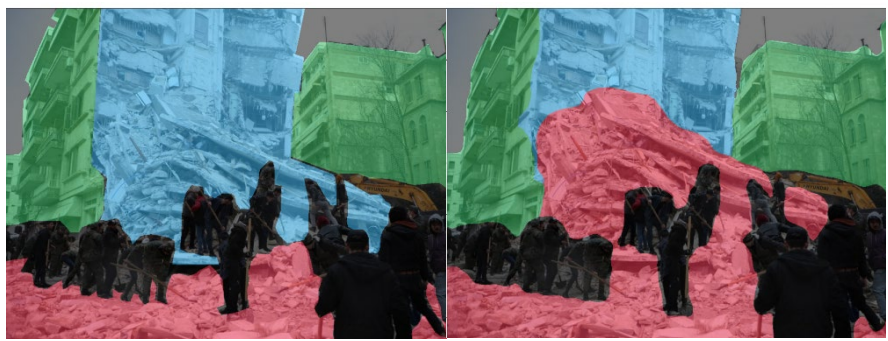

(a)                                             (b)

Fig. 3. Sample image pairs for IoU: (a) Annotation mask; (b) Segformer segmentation

## 2.2    SegFormer

SegFormer is a highly effective semantic segmentation model combining a Transformer-based hierarchical encoder with a multilayer perceptron (MLP) decoder, depicted in Figure 4 [21]. The SegFormer model has variants ranging from SegFormer-B0 to SegFormer-B5, each increasing in size and achieving progressively superior performance. This is particularly evident in benchmark datasets such as ADE20K [22] and Cityscapes [23], with SegFormer-B5 attaining an impressive 84% mean IoU on the Cityscapes validation set. The choice of SegFormer for this work stems from its proven efficacy in interpreting urban landscapes, as Cityscapes predominantly features urban elements like roads, sidewalks, buildings, walls, and fences [23]. Given that these elements are typically undamaged in the dataset, it's logical to adopt a Cityscapes-pretrained SegFormer model for assessing damage severity, ensuring relevance and accuracy in the context of this study.

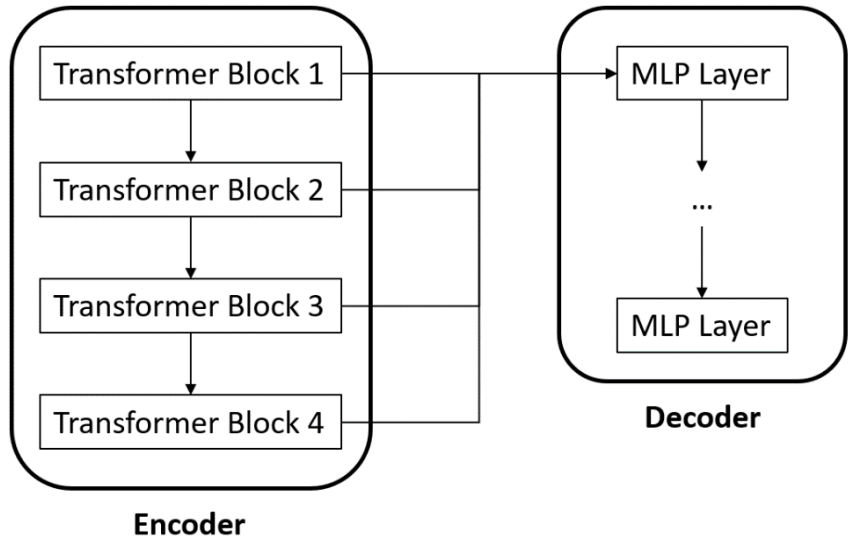

Fig. 4. SegFormer structure [21]

## 2.3 DPT

The assessment of damage severity in images encounters a fundamental dilemma: which represents more significant damage - a larger area with mild damage or a smaller area with severe damage [4]? This question underscores the significance of considering the area of damage in such assessments. However, direct measurement of area in images is challenging due to perspective distortion; objects closer to the camera appear larger, while those in the background appear smaller. Therefore, including depth information in the analysis is crucial. The consideration of depth allows for a more accurate interpretation of the actual area affected by damage, compensating for the apparent size differences caused by varying distances within the image.

In this work, the Dense Prediction Transformer (DPT) model is employed to generate depth estimates for each image [19]. This model integrates a vision transformer as its encoder, and a convolutional decoder. Notably, DPT provides relative depth estimations, instead of absolute depth measurements. Therefore, this approach considers depth on an intra-image basis, rather than inter-image comparisons. Given the absence of ground truth depth data for the images in this study, DPT is utilized in its pre-trained state for depth estimation, without any further fine-tuning.

### 2.4 Damage severity score

In this study, a scoring system is implemented to quantify the severity of damage depicted in images. As detailed in the DPT section, it is crucial to consider both the area affected and the degree of damage. The damage in an image is categorized into three degrees: undamaged structure (US), damaged structure (DS), and debris. The categorization is constrained to three degrees because the lower quality and limited observation angles typical of social media images pose challenges for more granular classification. Segmentation images generated by the finely-tuned SegFormer are used to calculate the area of each damage degree region. These areas are then adjusted using the depth estimation map provided by the DPT, ensuring a more accurate representation of the actual area affected by the various damage degrees.

A pixel-based damage severity equation is designed to calculate the damage severity score from the damage degree segmentation mask and the depth estimation.

Damage score =

$$\frac{\sum_i DS_{weight} \times DS_{pixel}(i) \times Depth_{pixel}(i) + \sum_j Debris_{pixel}(j) \times Depth_{pixel}(j)}{\sum_i DS_{pixel}(i) \times Depth_{pixel}(i) + \sum_j Debris_{pixel}(j) \times Depth_{pixel}(j) + \sum_m US_{pixel}(m) \times Depth_{pixel}(m)}$$

Eq. 1

Here, $i$ refers to the DS pixel, $j$ refers to the debris pixel and $m$ refers to the US pixel. The weight assigned to a DS pixel is set at 0.65, reflecting the relative severity of damage for damaged structure compared to that of debris. The value of the weight will be further explained in the Dataset section. $Depth_{pixel}$ is utilized to correct the area calculation according to depth information, assigning greater weight to pixels further in the background to compensate for perspective-induced size reduction. To be specific, the values of the depth estimation map are normalized to a range between 0.1 and 1 to determine the $Depth_{pixel}$ value. It is important to note that the chosen range starts at 0.1 rather than 0 to preserve the representation of damage severity in the pixels that are closest to the viewpoint. Background areas are excluded from damage severity calculations, as the focus is on the severity of damage itself in an image, irrespective of it in relation to the background. This approach ensures that the assessed damage severity remains consistent, whether viewed from a wider or zoomed in angle.

## 3 Dataset

To establish a ground truth for damage severity assessment, a dataset comprising 547 images is compiled using the search terms 'Turkey Earthquake', 'Wenchuan Earthquake', 'Haiti Earthquake', 'Nepal Earthquake', and 'Earthquake damage' on Google Search Engine. Images not related to post-disaster damage are removed from the dataset.

Two expert annotators are engaged to mask out US, DS, and debris from the images to mitigate subjectivity. The distinction between US and DS, and between DS and debris, presents certain ambiguities. Specific rules are applied during annotation to address the inherent ambiguities and maintain consistency between the two annotators:

1. Transition from US to DS: Structures visually without damage are marked as US. Note that in social media images, cracks may not be visually identifiable due to the relatively low resolution. Those structures with minor or structurally repairable damage have the damaged area labeled as DS and the rest as US; structures with extensive damage are entirely categorized as DS. However, if a lower section, such as the first floor, collapses but upper levels are intact, the upper levels will be classified as DS because the structural integrity of the building is compromised.

2. Transition from DS to debris: Visual recognition of a structure categorizes it as DS. When a structure is no longer identifiable, the resulting pile of material is considered debris. The entire area is labeled as debris in pancake collapse situations, where floors are compressed in layers.

Based on the annotation rules, if fitting US, DS and debris into Table 1, US will be within the damage severity range from 0 to 3, DS will be from 4 to 8 and debris will be from 9 to 10. Therefore, the weight of DS is chosen as 0.65 to represent the difference in damage severity between debris and DS.

While specific labeling rules are in place to guide the annotation process, the subjective nature of damage severity can lead to variations in interpretation. To reconcile these differences and achieve a unified final annotation, a final pass is done by the two annotators together in a conservative manner. In instances of disagreement or ambiguity, the annotation with a higher damage degree is chosen. This strategy ensures that the most conservative damage assessment is reflected in the final annotation. Given the focus on damage severity assessment and the potential goal of offering guidance to disaster reconnaissance team, adopting a more conservative standard is necessary, as it minimizes the risk of underestimating the extent and impact of the damage. Figure 5 shows five images in the dataset labeled by annotators 1 and 2, and the final annotation based on the conservative approach, where green refers to undamaged structure, blue refers to damaged structure, red refers to debris and black refers to background.

As the research progressed, the Morocco Earthquake in September 2023 provided an unforeseen opportunity to test the model's generalizability. A test dataset comprising 62 images was gathered using the same methodology previously employed, with the search term 'Morocco Earthquake'. This dataset is an ideal testing ground since the model developed in the Experiment section has not been exposed to data from the Morocco Earthquake. Utilizing this new dataset allows for a robust evaluation of the model's ability to generalize in the scenarios it has not previously encountered.

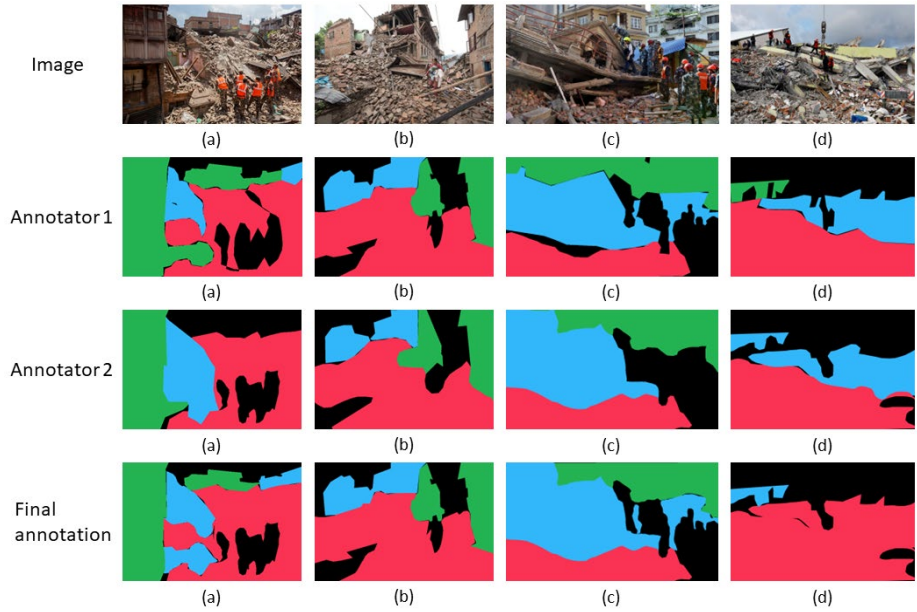

Fig. 5. Snapshot of the dataset and annotation (Green: undamaged structure; Blue: damaged structure; Red: debris; Black: background)

# 4 Experiment

## 4.1 Segformer Training and Validation for Damage Severity Assessment

The damage severity assessment dataset is divided into 80% for training and 20% for validation. A B5-sized SegFormer, pre-trained on CityScapes, is further fine-tuned on the damage severity assessment dataset to classify four classes: undamaged structure, damaged structure, debris, and background. Considering that the dataset is relatively small, early stopping is implemented to prevent overfitting by monitoring validation IoU. Table 2 shows the key configurations of the model.

Table 2. Segformer configuration

| Parameter | Value/Description |
|---|---|
| Batch size | 5 |
| Optimizer | AdamW |
| Learning rate | 0.00006 |
| Loss function | Cross entropy loss |
| Data augmentation | Horizontal flip and color jittering |
| Epoch | 100 |

The IoU achieved by the model on the validation dataset is 0.72. In the absence of pre-existing benchmarks or ground truth for the damage severity segmentation task, this performance is challenging to be directly assessed. To establish a reference standard, the IoUs between the annotators, and between each annotator and the final annotation are calculated, as illustrated in Figure 6. The IoU between annotator 1 and annotator 2 is 0.70, while the IoU scores of annotator 1 and annotator 2 with the final annotation are 0.74 and 0.75, respectively. These comparisons suggest that the model's performance, with an IoU of 0.72, is on par with human annotators, indicating its viability for automatic damage severity assessment.

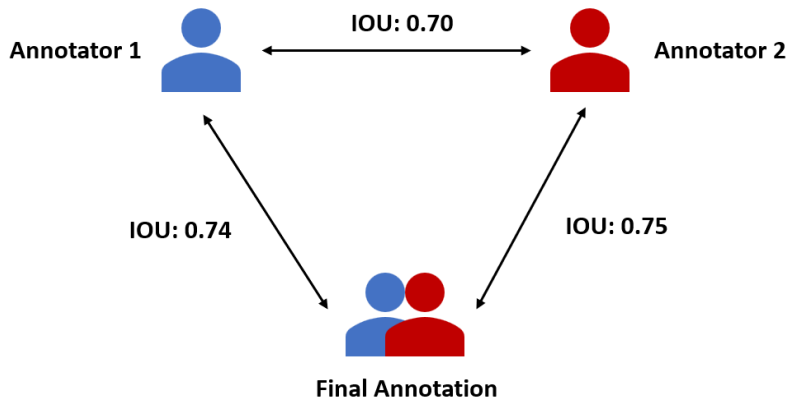

Fig. 6. IoU among human annotators

Some might contend that despite the model's IoU of 0.72 on the validation set, its generalizability could be limited due to the small dataset size. To address this concern, the fine-tuned model was additionally tested on the Morocco earthquake dataset, where it achieved an IoU of 0.63. This performance is noteworthy, especially considering that the model was not exposed to any images from the Morocco Earthquake during its training phase. The ability to maintain a reasonably high IoU on an entirely new dataset underscores the model's robustness and generalization capabilities in assessing damage severity.

### 4.2 Damage severity equation validation

To evaluate the damage severity equations, the EID dataset from a forthcoming paper serves as the benchmark [24]. The EID dataset comprises post-earthquake social media images categorized into four classes: three damage severity levels—little-to-no damage, mild damage, and severe damage—plus an irrelevant or non-informative class.

This experiment uses the EID dataset as ground truth to apply the damage severity assessment workflow (illustrated in Figure 1) to the images and determine if the calculated damage scores differ across the damage severity classes. From the EID dataset, 30 images are randomly selected for each of the little-to-no damage, mild damage, and severe damage classes. The fine-tuned Segformer, detailed in Section 4.1, generates damage segmentation, while DPT produces the depth image. Using these, the damage score is computed with Eq. 1. Figure 7 presents the average damage score for each class, showing a clear trend: as damage severity increases, the average damage score rises accordingly. This confirms that the damage severity equation effectively captures damage characteristics.

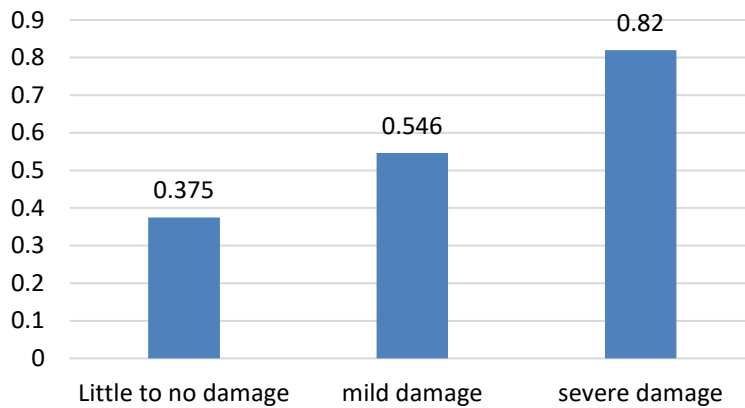

Fig. 7. Average damage severity score for the 90 images from EID dataset

## 5 Conclusion

This study advances the methodology for assessing earthquake damage severity from social media images, aiming to offer prompt and critical guidance to disaster rescue, recovery and reconnaissance teams. The shift from treating this challenge as a traditional classification problem to a semantic segmentation approach enriches the analytical depth and precision of damage assessments. By constructing a segmented dataset that categorizes damage into undamaged structures, damaged structures, and debris, this research integrates a finely-tuned SegFormer with the dataset, coupled with a novel damage severity scoring system. This system quantifies the degree of damage based on the area affected, adjusted by depth estimation, offering a nuanced view of damage severity that surpasses the subjective judgments of human annotators.

The introduction of the damage severity dataset marks a significant contribution to the field, providing future researchers with a valuable resource for training models on earthquake damage in social media images and establishing a baseline for this field of study. The proposed damage severity scoring system facilitates a more objective and detailed assessment of damage severity from social media images by considering the depth-adjusted area of different damage degrees within an image, rather than relying on direct human labeling. Moreover, this work underscores the potential of the Seg-Former model in the damage severity assessment field.

It's important to note that this study's scope is limited to assessing damage severity in post-disaster images on social media. It does not encompass the initial filtering of unrelated images. Future developments could address this aspect, aiming to create a pipeline capable of first isolating images pertinent to post-disaster damage and subsequently evaluating the damage severity through the method developed in this work. This would further streamline the process and augment the utility of social media imagery in disaster response and assessment.

# References

1. Bica, M., L. Palen, and C. Bopp. Visual representations of disaster. in Proceedings of the 2017 ACM conference on computer supported cooperative work and social computing. 2017.
2. Shekarizadeh, S., et al. Deep-Disaster: Unsupervised Disaster Detection and Localization Using Visual Data. in 2022 26th International Conference on Pattern Recognition (ICPR). 2022. IEEE.
3. Li, X., et al. Localizing and quantifying damage in social media images. in 2018 IEEE/ACM International Conference on Advances in Social Networks Analysis and Mining (ASONAM). 2018. IEEE.
4. Nguyen, D.T., et al. Damage assessment from social media imagery data during disasters. in Proceedings of the 2017 IEEE/ACM international conference on advances in social networks analysis and mining 2017. 2017.
5. Grünthal, G., European macroseismic scale 1998 (EMS-98). 1998.
6. American Red Cross. Red Cross/FEMA Disaster Assessment Pocket Guide. 2018 [cited 2024 Feb 06]; Available from: https://americanredcross.github.io/photo-survey/rc_fema_pocketguide.pdf.
7. FEMA. Hazus Earthquake Model Technical Manual 2.1. 2020 [cited 2024 Feb 06]; Available from: https://www.fema.gov/sites/default/files/2020-09/fema_hazus_earthquake-model_technical-manual_2.1.pdf.
8. Alam, F., F. Ofli, and M. Imran. Crisismmd: Multimodal twitter datasets from natural disasters. in Proceedings of the international AAAI conference on web and social media. 2018.
9. Alam, F., et al., MEDIC: a multi-task learning dataset for disaster image classification. Neural Computing and Applications, 2023. 35(3): p. 2609-2632.
10. Thoma, M., A survey of semantic segmentation. arXiv preprint arXiv:1602.06541, 2016.
11. Asad, M.H., et al. Natural Disaster Damage Assessment using Semantic Segmentation of UAV Imagery. in 2023 International Conference on Robotics and Automation in Industry (ICRAI). 2023. IEEE.

12. Chowdhury, T. and M. Rahnemoonfar. Attention based semantic segmentation on uav dataset for natural disaster damage assessment. in 2021 IEEE International Geoscience and Remote Sensing Symposium IGARSS. 2021. IEEE.

13. Doshi, J., S. Basu, and G. Pang, From satellite imagery to disaster insights. arXiv preprint arXiv:1812.07033, 2018.

14. Gupta, R., et al., xbd: A dataset for assessing building damage from satellite imagery. arXiv preprint arXiv:1911.09296, 2019.

15. Pi, Y., N.D. Nath, and A.H. Behzadan, Detection and semantic segmentation of disaster damage in UAV footage. Journal of Computing in Civil Engineering, 2021. 35(2): p. 04020063.

16. Rahnemoonfar, M., et al., Floodnet: A high resolution aerial imagery dataset for post flood scene understanding. IEEE Access, 2021. 9: p. 89644-89654.

17. Zhu, X., J. Liang, and A. Hauptmann. Msnet: A multilevel instance segmentation network for natural disaster damage assessment in aerial videos. in Proceedings of the IEEE/CVF winter conference on applications of computer vision. 2021.

18. Zhang, D.Y., et al. Crowd-assisted disaster scene assessment with human-ai interactive attention. in Proceedings of the AAAI Conference on Artificial Intelligence. 2020.

19. Ranftl, R., A. Bochkovskiy, and V. Koltun. Vision transformers for dense prediction. in Proceedings of the IEEE/CVF international conference on computer vision. 2021.

20. Lee, S.-H. and H.-C. Chen, U-SSD: Improved SSD based on U-Net architecture for end-to-end table detection in document images. Applied Sciences, 2021. 11(23): p. 11446.

21. Xie, E., et al., SegFormer: Simple and efficient design for semantic segmentation with transformers. Advances in Neural Information Processing Systems, 2021. 34: p. 12077-12090.

22. Zhou, B., et al., Semantic understanding of scenes through the ade20k dataset. International Journal of Computer Vision, 2019. 127: p. 302-321.

23. Cordts, M., et al. The cityscapes dataset for semantic urban scene understanding. in Proceedings of the IEEE conference on computer vision and pattern recognition. 2016.

24. Huang, H. et al., Enhancing the Fidelity of Social Media Image Datasets in Earthquake Damage Assessment, Earthquake Spectra, 2025 (in press).